\documentclass[runningheads]{llncs}
\usepackage{geometry}
\usepackage{graphicx}

\usepackage{tikz}
\usepackage{comment}
\usepackage{amsmath,amssymb} 
\usepackage{color}
\usepackage{booktabs}
\usepackage{caption}
\usepackage{subcaption}
\usepackage{multirow}
\usepackage{colortbl} 
\usepackage{array} 
\usepackage{hhline}
\usepackage{algorithm}
\usepackage{algorithmic}
\usepackage[]{hyperref} 
\usepackage{float}

\makeatletter
\renewcommand*{\@fnsymbol}[1]{\ensuremath{\ifcase#1\or *\or \dagger\or \dagger\dagger\or
    \mathsection\or \mathparagraph\or \|\or **\or \ddagger
    \or \ddagger\ddagger \else\@ctrerr\fi}}
\makeatother

\begin{document}
\pagestyle{headings}
\mainmatter
\def\ECCVSubNumber{100}  

\title{Meta-Regularization by Enforcing Mutual-Exclusiveness} 


\author{Edwin Pan\thanks{Equal contribution} \and Pankaj Rajak\printfnsymbol{1}\thanks{Author is also affiliated with Argonne National Laboratory, Lemont, IL [prajak@anl.gov]} \and Shubham Shrivastava\printfnsymbol{1}\thanks{Author is also affiliated with Ford Greenfield Labs, Palo Alto, CA [sshriva5@ford.com]}\\
\footnotesize\{edwinpan, prajak7, shubhams\}@stanford.edu}
\institute{Stanford University}

\authorrunning{Edwin Pan \and Pankaj Rajak \and Shubham Shrivastava}

\maketitle

\begin{abstract}
Meta-learning models have two objectives. First, they need to be able to make predictions over a range of task distributions while utilizing only a small amount of training data. Second, they also need to adapt to new novel unseen tasks at meta-test time again by using only a small amount of training data from that task. It is the second objective where meta-learning models fail for non-mutually exclusive tasks due to task overfitting. Given that guaranteeing mutually exclusive tasks is often difficult, there is a significant need for regularization methods that can help reduce the impact of task-memorization in meta-learning. For example, in the case of N-way, K-shot classification problems, tasks becomes non-mutually exclusive when the labels associated with each task is fixed. Under this design, the model will simply memorize the class labels of all the training tasks, and thus will fail to recognize a new task (class) at meta-test time. A direct observable consequence of this memorization is that the meta-learning model simply ignores the task-specific training data in favor of directly classifying based on the test-data input. In our work, we propose a regularization technique for meta-learning models that gives the model designer more control over the information flow during meta-training. Our method consists of a regularization function that is constructed by maximizing the distance between task-summary statistics, $h(D_i^{tri})$, in the case of black-box models and task specific network parameters $(\phi_{i})$ in the case of optimization based models during meta-training. Maximizing the proposed regularization function reduces task-overfitting because both  $h(D_i^{tri})$ and $(\phi_{i})$ are computed using task training data $(D_i^{tri})$, and thus encourages the model to use $D_i^{tri}$ during meta-training. Our proposed regularization function shows an accuracy boost of $\sim$ $36\%$ on the Omniglot dataset for 5-way, 1-shot classification using black-box method and for 20-way, 1-shot classification problem using optimization-based method. \footnote{Code implementation available here: \url{https://github.com/towardsautonomy/meta_reg_by_enforcing_mutual_exclusiveness} } 

\end{abstract}

\section{Introduction} 
Designing machine learning (ML) algorithms that are sample-efficient and generalize to novel tasks is a challenging problem and an active area of research. ML models based on supervised learning have shown tremendous success in solving many complex problems across various domains. However, these models also require large amounts of data and have lower performance on datasets that are significantly different from the training data distribution. In recent years, meta-learning has emerged as a popular framework to design models that are robust, sample-efficient, have high accuracy over a wide range of task distributions, and also adapt to new unseen tasks with only few training examples. \\

Even though meta-learning has shown exciting results over a range of problems, they require the careful construction of the input training task distribution. Without this careful construction, they suffer from a problem known as task-overfitting, which is different from the issue of overfitting associated with the supervised learning models. Overfitting in supervised learning happens when the model memorizes the input training data, and exhibits low performance on the test dataset samples which are drawn from the same distribution of training data. In contrast, task-overfitting in meta-learning happens because the model memorizes all the training tasks, and thus have low performance on new unseen tasks at meta-test time. For this reason, known regularization techniques such as dropout, data-augmentation and $L_2$ regularization of network weights do not prevent task-overfitting. All these techniques work at the dataset level for single task problems. Thus, novel regularization techniques need to be designed to work on over task-distribution in meta-learning. \\

In all meta-learning models, each task consists of training and test data, where the objective of the model is to quickly converge on a solution for the task using task-training data only. However, during training if models completely ignore the task-training data while learning about the tasks, it leads to task-overfitting. In essence, the model has memorized the underlying functional form of all the training tasks in its weight vector, and thus does not know how to process the training data of new tasks at meta-test time. Thus, the objective of all meta-regularization techniques is based on the principal that meta-learning does not ignore task training data while learning about the tasks. \\
 
The primary contribution of our work consists of:
\begin{enumerate}
    \item Identification and analysis of task-memorization in black-box and optimization based meta learning models.
    \item Construction of a meta-regularization function that forces the use of task-training data while learning about the tasks. In black-box models, we do that by maximizing the Euclidean distance between the summary statistics of tasks that are computed using the task-training data. In the case of optimization based models, we regularize the model by maximizing the Euclidean distance between the task-specific parameters itself. 
\end{enumerate}


\section{Related Work} 
Meta-learning enables the ability to learn new tasks with small number of available data points quickly by utilizing experiences from previous related tasks \cite{thrun_1998}, \cite{287172}, \cite{pmlr-v48-santoro16}, \cite{finn2017modelagnostic}, \cite{nichol2018firstorder}. These methods have shown encouraging results, however, recent papers like \cite{zhang+al-2018-metagan}, \cite{yin2020metalearning}, and \cite{lee2020meta} have established and formalized the meta-overfitting problem with meta-learning. Previous works have approached the issue of meta-overfitting in different ways.

Yin et al. \cite{yin2020metalearning} proposed a regularization technique based on maximizing the mutual information between the task-training data and task-specific parameters. They do this by drawing the meta-parameter $\theta$ from a Gaussian distribution instead of point estimate value. 

Lee et al. \cite{lee2020meta} on the other hand proposed meta-dropout, which uses a separate model to learn the noise distribution to perturb each dimension of the input training and test data such that it becomes harder for the model to memorize all the tasks. 

Rajendran et al. \cite{rajendran2020metalearning} proposed a slightly different approach, where they introduced randomness into the data, thereby discouraging the model from learning trivial solutions. They show that meta-augmentation can improve the performance of base model in adapting to a new dataset.

Our approach towards tackling the memorization problem in meta-learning is simple and intuitive. We try to enforce mutual-exclusiveness between tasks by maximizing the distance between task-specific model parameters. This discourages our model from learning a base model which can infer tasks from test input alone and thus our model does not collapse to a zero-shot solution. This is more clearly explained in section \ref{sec:method}.


\section{Memorization in Meta-Learning} 

The meta-learning problem statement is often framed in contrast to the mathematical framework behind the classical supervised machine learning problem. Specifically, the standard supervised machine learning problem uses training data and label pairs $\mathcal{D} = (x_i, y_i)$, sampled i.i.d. from a single task $\mathcal{T}$, to learn a function $x \mapsto \hat{y}$. Similarly, in meta-learning our goal is to \textit{learn-to-learn} by sampling a set of tasks $\{\mathcal{T}_i\}$ i.i.d., where each task is composed of both a support set of training data-label pairs $\mathcal{D}_i^{tri} = (x_i^{tri}, y_i^{tri})$ and a query set of training data-label pairs $\mathcal{D}_i^{test} = (x_i^{test}, y_i^{test})$. Thus, at meta-test time, our goal is to quickly learn a model for a new task $\mathcal{T}$ using only a small sample of data along with the learned meta-parameters $\theta$ and task-specific-parameters $\phi$. The entire set of meta-training tasks is denoted by $\mathcal{M} = \{ \mathcal{D}_i^{tri}, \mathcal{D}_i^{test}\}_{i=1}^{N}$. \\
\begin{figure}[!htbp]
\centering
    \begin{subfigure}{0.30\linewidth}
    \centering
    \includegraphics[width=\linewidth]{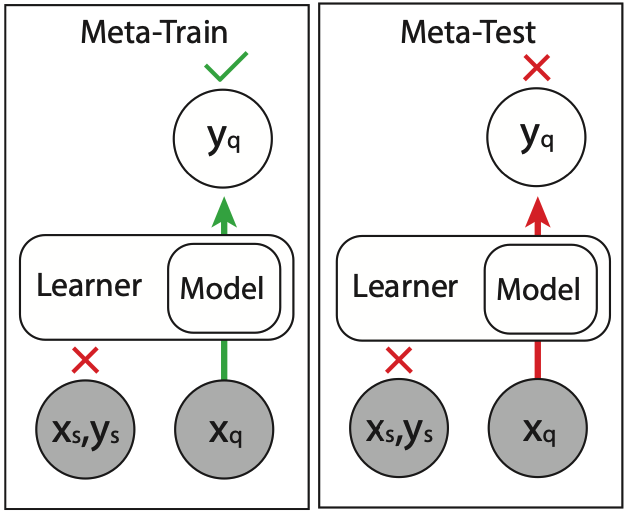}
    \caption{Memorization overfitting occurs when the meta-learner ignores training examples and directly infers from a given query task \cite{rajendran2020metalearning}}\label{fig:mem_overview}
    \end{subfigure}
    \hspace{0.1in}
    \begin{subfigure}{0.60\linewidth}
    \centering
    \includegraphics[width=\linewidth]{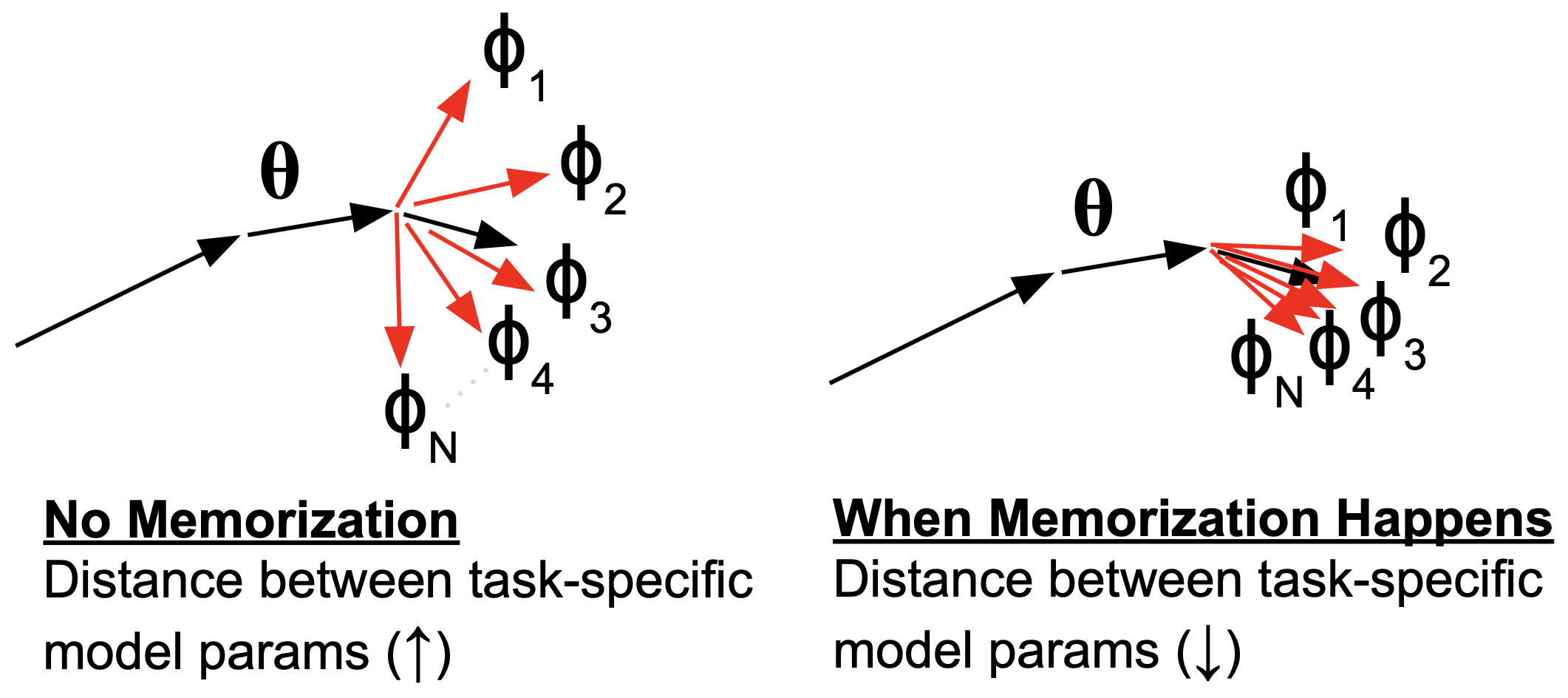}
    \caption{In MAML, the task-specific parameters tend to get very close to each other when memorization happens.}\label{fig:mem_maml_params}
    \end{subfigure}
    \caption{Memorization in Meta-Learning}\label{fig:memmorization_in_meta_learning}
\end{figure}

Designing a set of mutually-exclusive tasks takes a tremendous amount of work as it's non-trivial and highly problem specific. One example is the technique of shuffling the labels in a classification problem. By making the class labels inconsistent throughout the training process, we ensure that the model cannot directly memorize the labels associated with each class. Unfortunately, the same strategy of label-shuffling does not work for all types of problems (ex: label-shuffling doesn't work with regression). For this reason, it's not always possible to construct mutually exclusive tasks from a training-task distribution. When tasks are non-mutually exclusive, the simplest way for the model to solve the given problem is to directly learn a single function that can solve all the given tasks, which causes task-memorization. In other words, the meta-learner puts less emphasis on task-specific training data and attempts to accurately infer outcomes based more on the test input. Since a desired meta-learner has to utilize both sources of information, each meta-learning model exists on a spectrum based on its weighting of task-specific information and meta information. Consequently, memorization only becomes an issue when the meta-learning model places so much emphasis on the meta information that the model become incapable of generalizing to unseen tasks. At the limit of complete memorization, the model will only draw conclusions based on the test input (Figure \ref{fig:mem_overview}). How the memorization problem manifests depends on the meta-learning algorithm used, and the model architecture implemented. In this paper, we explore memorization in Memory-Augmented Neural Networks (MANN) \cite{pmlr-v48-santoro16} and Model-Agnostic Meta-Learning (MAML) \cite{finn2017modelagnostic}, two popular meta-learning algorithms. \\

As we mentioned above, meta-learning models need to learn based on both task-specific information and meta-information across tasks. In the context of MANN based meta-learning (sometimes referred to as a "black-box" based approach to meta-learning), task-specific information and meta information can be observed through the \textit{task-summary} statistics and the \textit{latent representation of tasks}, respectively. The quantities are defined in detail below. \\

\begin{figure}[!htbp]
\centering
    \centering
    \includegraphics[width=1.0\linewidth]{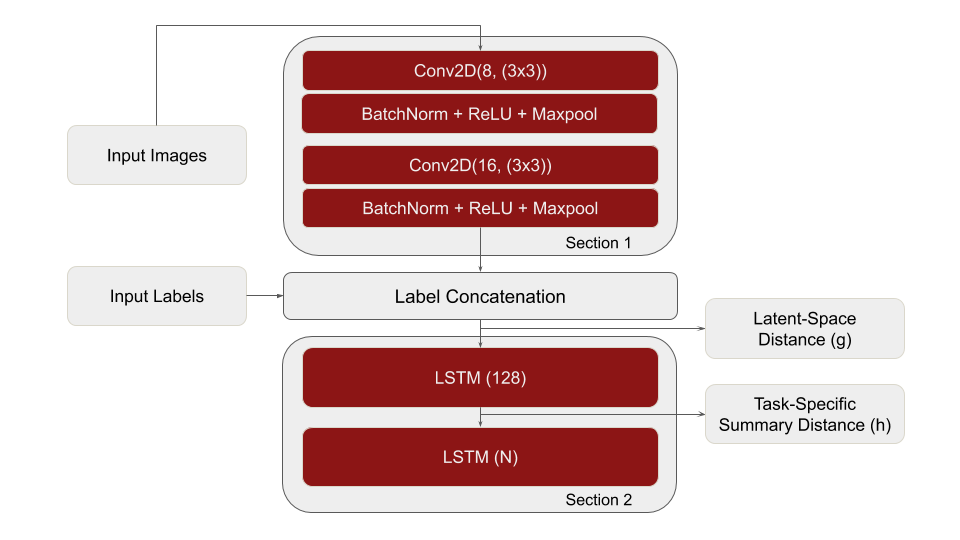}
\caption{MANN Model Architecture.}\label{fig:MANN_architecture}
\end{figure}

The MANN model architecture used in these experiments is composed of an encoder to learn latent representation $(g_\theta)$ of input data, and a LSTM block \cite{HochSchm97} as a black-box meta-learning model. The encoder consists of a series of convolutional neural networks \cite{NIPS2012_c399862d}. After computing the latent representation $(g_\theta)$ of the training data $D^{tri}_i$ for a given task, we concatenate each $g_\theta$ with their corresponding labels, which is then feed into the LSTM block to compute task-summary statistics $(h_\theta(D^{tri}_i))$. This $h_\theta(D^{tri}_i)$ is then used to learn a task-specific function and make predictions on task-test data $(D^{test}_i)$. Details of the architecture used are included in Figure \ref{fig:MANN_architecture}. To regularize the model such that $D^{tri}_i$ does not get ignored, we  maximized the distance between the summary statistics of tasks as $f_{distance}(h_{\theta}(D_i^{tri}), h_{\theta}(D_j^{tri}))$. Further, since the distance between $D^{tri}_i$ for similar tasks should be smaller, and dissimilar tasks should be larger, we have also used another regularization function where $f_{distance}(h_{\theta}(D_i^{tri}), h_{\theta}(D_j^{tri}))$ is normalized by the distance between task latent-representation $f_{distance}(g_{\theta}(D_i^{tri}), g_{\theta}(D_j^{tri}))$. Here, we have used L2-norm as distance metric function $f_{distance}()$. \\
 
In the black-box model context, we treat $h_\theta$ as analogous to stepping through a model with task-specific parameters $\phi$, and $g$ as analogous to stepping through a model with meta-parameters $\theta$. Therefore, during memorization we expect to observe a high value for latent-representation distance $(f_{distance}(g_{\theta}(D_i^{tri}), g_{\theta}(D_j^{tri})))$ relative to the task-summary distance $(f_{distance}(h_{\theta}(D_i^{tri}), h_{\theta}(D_j^{tri})))$. \\

For MAML based meta-learning, memorization results in the the model meta-parameter $\theta$ getting very close to the task-specific parameters $\phi_i; i \in \textbf{m}$, where \textbf{m} is the number of tasks within the current batch. This means that the model starts ignoring \textit{meta-training} data and does not learn distinct $\phi_i$ for each task. When this happens, the distance between task-specific parameters gets very low; i.e. $\phi_i \approx \phi_j \forall i \neq j$. Initialization model meta-parameter $\theta$ thus learnt is not good enough to adapt to a new task quickly and hence, we see a very low validation accuracy on \textit{meta-validation}. This problem can be alleviated to some extent by enforcing the task-specific parameters $\phi_i$ to be far from $\phi_j$ in some proportion to the distance between $task_i$ and $task_j$ themselves, $\forall i \neq j$.

\section{Our Method - Mutual Exclusiveness Regularization}\label{sec:method} 
The main contribution of this paper is a proposed method of regularization to reduce the impact of memorization, as well as experiments to validate and analyze the performance of said regularization method.

\subsection{Black-Box Method (MANN)}
The general MANN network objective is:
\begin{equation}
    \min_{\theta} \sum_i (\mathcal{L}(h(D_i^{tri}), D_j^{test})
\end{equation}

For the MANN model experiments, we explore three different methods of regularization. As noted in the previous section, memorization in MANN is observed when the ratio of task-summary distance to latent-representation distance is low.  In method 1, we aim to improve this ratio by adding a regularization incentive for larger values of task-summary distance. Mathematically, this is done by training with the following objective.
\begin{equation}
    \min_{\theta} \sum_i (\mathcal{L}(h(D_i^{tri}), D_j^{test})-\lambda \sum_{i}\sum_{j,j\neq i} f_{distance}(h_{\theta}(D_i^{tri}), h_{\theta}(D_j^{tri}))
\end{equation}

Not all task-summary distances should be weighted equally. In fact, if two tasks are similar, we would want the model to focus more on the task-summary data. Therefore, the second method of regularization weights the task-summary distances by the respective task-latent distance.
\begin{equation}
    \min_{\theta} \sum_i (\mathcal{L}(h(D_i^{tri}), D_j^{test})-\lambda \sum_{i}\sum_{j,j\neq i} \frac{f_{distance}(h_{\theta}(D_i^{tri}), h_{\theta}(D_j^{tri}))}{f_{distance}(g(D_i^{tri}), g(D_j^{tri}))})
\end{equation}

The end goal of regularizing the objective function used by a meta-learning model is to give us better control over the information flow. The final method of regularization defines a desired ratio of task-summary information to task-latent information $\eta$. While this does introduce another hyper-parameter that would need to be tuned, it allows us much greater control over the model's behavior. 
\begin{equation}
    \min_{\theta} \sum_i (\mathcal{L}(h(D_i^{tri}), D_j^{test})+\lambda \sum_{i}\sum_{j,j\neq i} |\eta - \frac{f_{distance}(h_{\theta}(D_i^{tri}), h_{\theta}(D_j^{tri}))}{f_{distance}(g(D_i^{tri}), g(D_j^{tri}))}|)
\end{equation}

\subsection{Optimization-Based Method (MAML)}

In an optimization-based method like MAML, the network parameters are optimized for each task in an inner loop, and later the parameters are optimized to minimize loss over all the tasks. In the process, the network learns a good initialization, from which it can solve a new task in a very few number of inner updates. If the task-specific model parameters used during inner loop updates are denoted as $\phi$ and the initialization parameters are denoted as $\theta$, then the memorization problem in MAML can be summarized as shown in Fig \ref{fig:mem_maml_params}. \\

\begin{figure}[!t]
\centering
    \centering
    \includegraphics[width=0.99\linewidth]{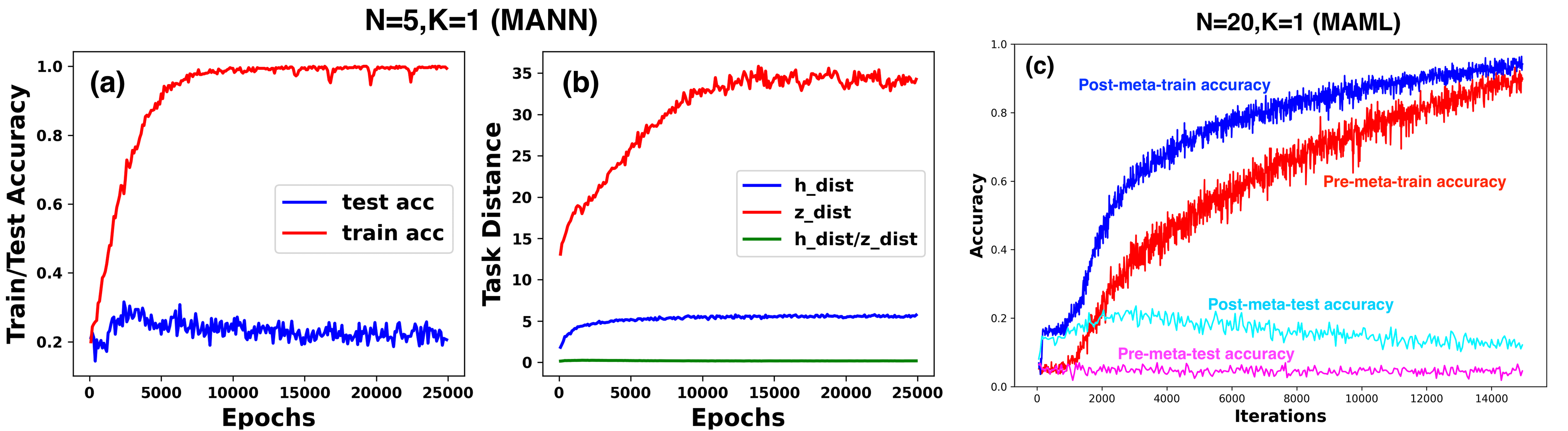}
\caption{Memorization in (a-b) Black-box (MANN) and (c) Optimization-based(MAML) model. In MANN, when memorization happens, distance between task-summary statistics($h_{dist}$) gets very low which means the model has ignored the task training data. (c) Shows memorization in optimization-based (MAML) model, where memorization happens the task specific parameters $\phi$ gets very close to the model parameters $\theta$. This results in the pre-optimization model accuracy getting very close to the post-optimization accuracy, and thus the network performance is very poor at validation/test time.}\label{fig:fig3-ver2}
\end{figure}

If all the tasks are mutually-exclusive, the model parameters learnt during the inner loop are far from each other. However, if the tasks are not mutually-exclusive, as the model starts to remember training samples, initialization parameters learnt during the training process starts to solve all of the tasks. As a result, task parameters $\phi_i \approx \phi_j; \forall i \neq j$, which also means that $\theta \approx \phi_i; \forall i$.  This can be clearly seen in Fig \ref{fig:fig3-ver2}c, where the pre-optimization accuracy increases as the training progresses. For mutually-exclusive tasks, the pre-optimization accuracy should correspond to a random guess while the post-optimization accuracy approaches $1.0$ during training. Since the model memorizes training samples during \textit{meta-training} step, the post-optimization accuracy during the \textit{meta-validation} step remains quite low as shown in Fig \ref{fig:fig3-ver2}c. \\

\begin{figure}[!htbp]
\centering
    \centering
    \includegraphics[width=0.55\linewidth]{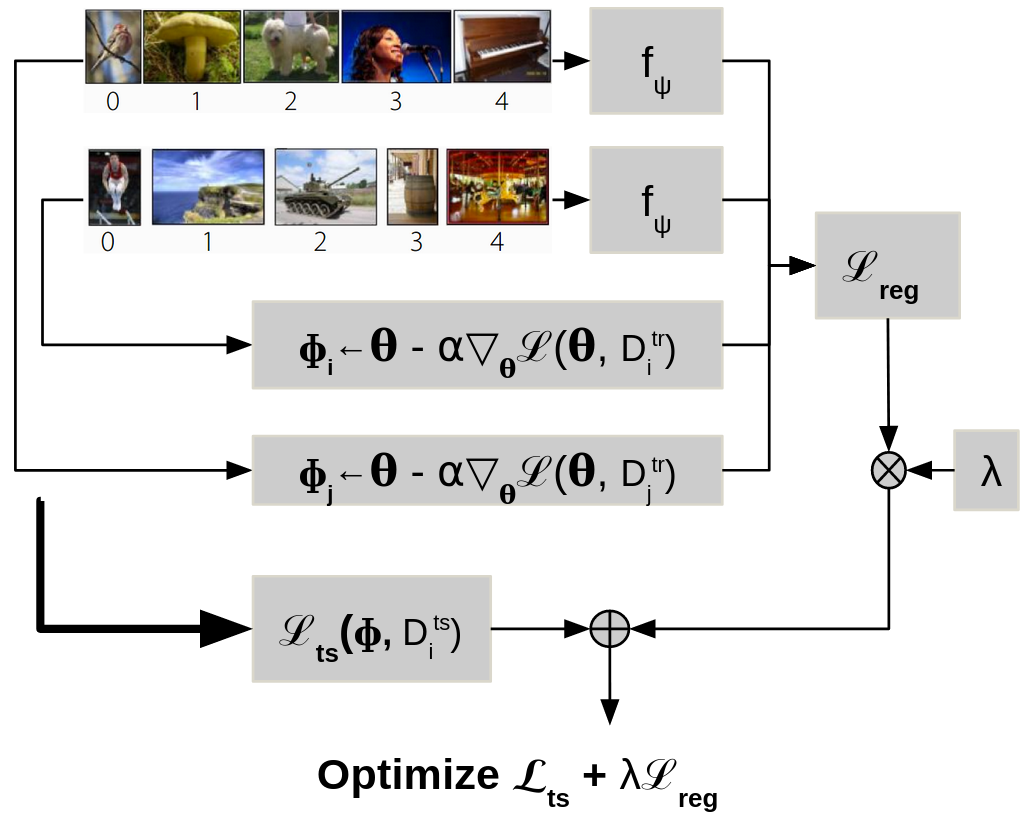}
\caption{Our meta-regularization approach for optimization-based meta learning.}\label{fig:metareg_block_diagram}
\end{figure}

We formulate our regularization function to target the memorization problem by enforcing \textit{mutual-exclusiveness} in a set of \textit{non-mutually-exclusive} tasks. We do this by enforcing the task-specific parameters $\phi_i$ to be farther away from $\phi_j$. As shown in Fig \ref{fig:metareg_block_diagram}, we first perform inner loop updates for all tasks to obtain $\phi_i; i \in [0, \textbf{m})$, where \textbf{$m$} is the total number of tasks. During the outer loop updates, the regularization loss function, $\mathcal{L}_{reg}$ is obtained by sampling \textit{$k$} pairs of tasks and computing a \textit{negative} measure of distance between task-specific model parameters. This is then weighted by a factor $\lambda$ and then added to the outer-loop loss function $\mathcal{L}_{ts}$ to obtain a total loss function $\mathcal{L} = \mathcal{L}_{ts} + \lambda \mathcal{L}_{reg}$. Additionally, we pose the constraints that the difference between task-specific parameters should be proportional to the distance between the tasks themselves, obtained by extracting latent embeddings of tasks. We denote this proportionality constant by $\eta$. The complete algorithm for MAML with Mutual Exclusiveness Regularization (\textit{MER-MAML}) is given in Algorithm \ref{alg:maml_mer}. \\

\newcommand{\task}{\mathcal{T}}
\newcommand{\loss}{\mathcal{L}}
\newcommand{\inp}{\mathbf{x}}
\newcommand{\learner}{f}
\newcommand{\lossi}{\loss_{\task_i}}

\begin{algorithm}[!ht]
\caption{Model-Agnostic Meta-Learning with Mutual Exclusiveness Regularization}
\label{alg:maml_mer}
\begin{algorithmic}[1]
\REQUIRE $p(\task)$: distribution over tasks
\REQUIRE $\alpha$, $\beta$: step size; $\lambda$: regularization factor; $\eta$: task parameters distance to task embedding distance ratio; $K$: number of task pairs to sample, $K < $ number of tasks in current batch
\STATE randomly initialize $\theta$
\WHILE{not done}
\STATE Sample batch of tasks $\task_i \sim p(\task)$
  \FORALL{$\task_i$}
 \STATE Evaluate $\nabla_\theta \lossi(\learner_\theta)$ with respect to $K$ examples
 \STATE Compute task specific parameter with gradient descent: $\phi_i=\theta-\alpha \nabla_\theta  \lossi(  \learner_\theta )$
 \STATE Obtain task encoding: $f(D_i^{tr})$
 \STATE Obtain task-specific model parameters: $g(f(D_i^{tr}))$
 \ENDFOR
 \FOR{\textit{k}=1:K}
  \STATE Sample pair of tasks: {$\task_i$, $\task_j$}
  \STATE Compute distance between task encodings: $f_{distance}(f(D_i^{tr}), f(D_j^{tr}))$
  \STATE Compute distance between task-specific model parameters: $f_{distance}(\phi_i, \phi_j)$
  \STATE Compute: $\mathcal{R}_{{\phi}_k} = \eta - \frac{f_{distance}(\phi_i, \phi_j)}{f_{distance}(f(D_i^{tr}), f(D_j^{tr}))}$
 \ENDFOR
 \STATE Compute: $\mathcal{R}_{\phi} = \frac{1}{K}\sum_k \mathcal{R}_{{\phi}_k}$
 \STATE Update $\theta \leftarrow \theta - \beta [\nabla_\theta \sum_{\task_i \sim p(\task)}  \lossi ( \learner_{\phi_i}) + \lambda \mathcal{R}_{\phi}]$
\ENDWHILE
\end{algorithmic}
\end{algorithm}

In the algorithm outlined (Algorithm \ref{alg:maml_mer}), $\alpha$ is the inner-loop learning rate and $\beta$ is the meta-step size used during stochastic gradient descent (SGD). The complete meta-objective function is given in Equation \ref{eq:MAML_obj_eq}.

\begin{equation}\label{eq:MAML_obj_eq}
\min_\theta \sum_{\task_i \sim p(\task)}  \lossi ( \learner_{\phi_i}) + \lambda [\eta - \frac{f_{distance}(\phi_i, \phi_j)}{f_{distance}(g(D_i^{tr}), g(D_j^{tr}))}]
\end{equation}

\section{Experimental Details} 

\subsection{MER-MANN}
The MANN model used in the experiments below was trained and tested on the Omniglot dataset \cite{Lake_oneshot}, with 1200 training examples, 400 validation examples, and 23 test examples. In prior sections, we noted that memorization can be alleviated by clever task construction to ensure mutual exclusivity. For Omniglot dataset image classification, mutual exclusivity can be introduced by assigning shuffled labels to various images. This way, the model will see multiple different labels during training time with no way for the model to use that information to directly arrive at a solution. In our experiments, we induce memorization by \textit{intentionally} assigning consistent labels, observing memorization behavior, then using our regularization methods to reduce the behavior. The experiments include both $5$-\textit{way} $1$-\textit{shot} classification and $10$-\textit{way} $1$-\textit{shot} classification. For the third method of regularization, we also perform parameter search in order to find the ideal $\eta$ that maximizes the test accuracy. \\

Training is done with batch size of 32. The convolutional networks and LSTMs are optimized using Adam with a learning rate of 0.001. For each method of regularization defined above, we used $\lambda=1$.

\subsection{MER-MAML}

\begin{figure}[!hb]
\centering
    \begin{subfigure}{0.3\linewidth}
    \centering
    \includegraphics[width=\linewidth]{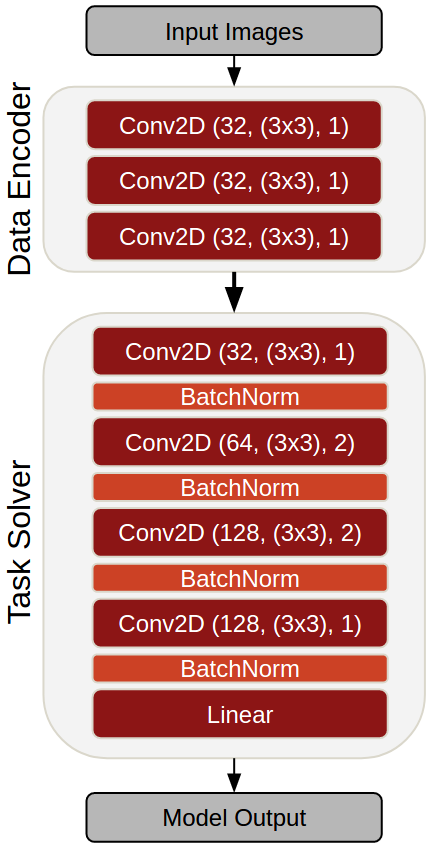}
    \caption{Model architecture used for MER-MAML, \textit{Conv2D} layers are represented as $Conv2D(filters, (k x k), stride)$}\label{fig:mer_maml_model_architecture}
    \end{subfigure}
    \hspace{0.15in}
    \begin{subfigure}{0.6\linewidth}
    \centering
    \vspace{0.35in}
    \includegraphics[width=\linewidth]{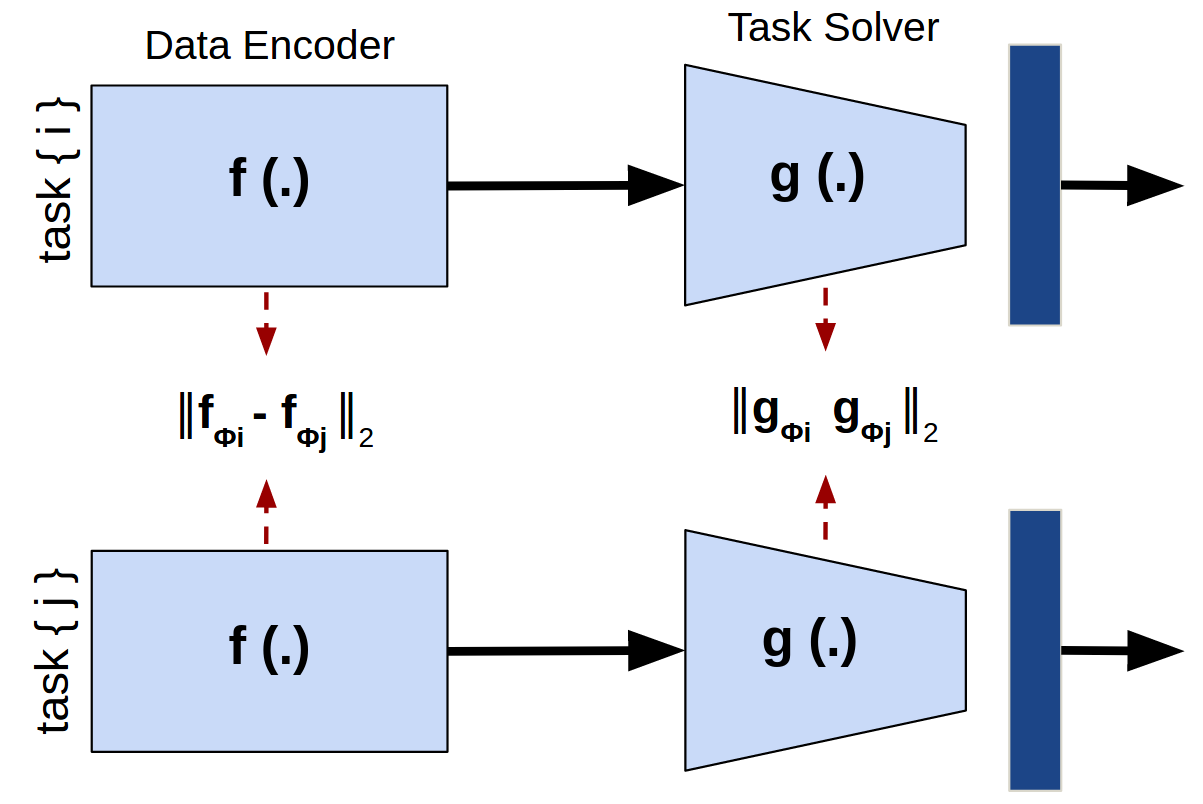}
    \vspace{0.3in}
    \caption{Our proposed regularization method for tackling memorization in MAML - \textit{Meta Exclusiveness Regularization (MER)}}\label{fig:mer_maml_block_diagram}
    \end{subfigure}
    \caption{Model architecture and our meta exclusive regularization method for MAML}\label{fig:mer_maml}
\end{figure}

Obtaining a task representation for measuring the distance between task embeddings is tricky. One can use a pre-trained feature extractor to obtain the latent vector and use a separate function to measure the distance between two such vectors. In our case, we construct our model as a combination of \textit{Data Encoder} and \textit{Task Solver}. While the former is responsible for extracting task data-point features, the latter is in charge of solving each task. In order to obtain distances between task embeddings, we first extract \textit{Data Encoder} model parameters for each task and then use a distance metric (\textit{L2} norm in our experiments) to compute the distance between a pair of tasks. Similarly, to compute distances between task-specific parameters, we obtain distances between \textit{Task Solver} model parameters from the same pair of tasks. This is demonstrated in Fig \ref{fig:mer_maml}. \\

For all our experiments with \textit{MER-MAML}, we use inner loop learning rate of $0.04$ and outer loop learning rate of $0.0025$. The number of sampled task pairs was chosen to be $4$. All the experiments listed in Table \ref{tab:ablation_mer_maml} were conducted with a batch size of $10$. We work with the Omniglot dataset \cite{Lake_oneshot} for solving a $20$-\textit{way}, $1$-\textit{shot}, classification problem. The Omniglot dataset contains 20 instances of $1623$ characters from 50 different alphabets. We divide these characters into $1100$ \textit{meta-training}, $100$ \textit{meta-validation}, and $423$ \textit{meta-testing} set. To maintain the label consistency throughout the experiments, we divide the sets into fixed $n$-\textit{classes} for $n$-\textit{way} classification problem. We also conducted some preliminary experiments with the Pose dataset \cite{xiang_wacv14} for pose regression task.

\section{Results} 

\subsection{MER-MANN}
In MANN based meta-learning, we observe memorization when training accuracy increases greatly while test accuracy remains low (see Figure \ref{fig:fig3-ver2}a). We experiment with various regularization methods all under the umbrella designation MER-MANN. We find that our proposed MER-MANN model improves test accuracy, though at great man-hour costs in tuning parameters to find the right emphasis ratio needed for the desired $\mathcal{M}$. Results for the experiments detailed above are reported in Table \ref{tbl:mer_mann_results}.

\begin{table}[!hbt]\footnotesize 
\begin{center}
\caption{Validation accuracy of MANN models with our regularization, \textit{Meta Exclusiveness Regularization (MER)}, on Omniglot datasets}
\label{tbl:mer_mann_results}
\begin{tabular}{|l|*{4}{m{2.5cm}|}m{2.5cm}} \hline
 & \multicolumn{2}{c|}{\textbf{\textit{5}}-way \textbf{\textit{1}}-shot Classification}  & \multicolumn{2}{c|}{\textbf{\textit{10}}-way  \textbf{\textit{1}}-shot Classification} \\ \hhline{>{\arrayrulecolor{white}}-*{4}{>{\arrayrulecolor{black}}-}}
\multirow{-2}{*}{}& Test Accuracy \% & $h_{\theta} / z_{\theta}$ Ratio & Test Accuracy \% & $h_{\theta} / z_{\theta}$ Ratio \\ \hline
 &  & & & \\
\multirow{-2}{*}{Label Shuffled MANN} & \multirow{-2}{*}{45}& \multirow{-2}{*}{--}& \multirow{-2}{*}{32}& \multirow{-2}{*}{--}\\ \hline
 &  & & & \\
\multirow{-2}{*}{Memorization MANN} & \multirow{-2}{*}{23}& \multirow{-2}{*}{0.14}& \multirow{-2}{*}{11}& \multirow{-2}{*}{0.1}\\ \hline
 &  & & & \\
\multirow{-2}{*}{MER Method 1} & \multirow{-2}{*}{41}& \multirow{-2}{*}{3.2}& \multirow{-2}{*}{12}& \multirow{-2}{*}{1.3}\\ \hline
 &  & & & \\
\multirow{-2}{*}{MER Method 2} & \multirow{-2}{*}{45}& \multirow{-2}{*}{2.3}& \multirow{-2}{*}{10.6}& \multirow{-2}{*}{0.01}\\ \hline
 &  & & & \\
\multirow{-2}{*}{MER Method 3} & \multirow{-2}{*}{59}& \multirow{-2}{*}{2.0}& \multirow{-2}{*}{26}& \multirow{-2}{*}{4.0}\\ \hline

\end{tabular}
\end{center}
\end{table}

\begin{figure}[!t]
\centering
    \centering
    \includegraphics[width=0.9\linewidth]{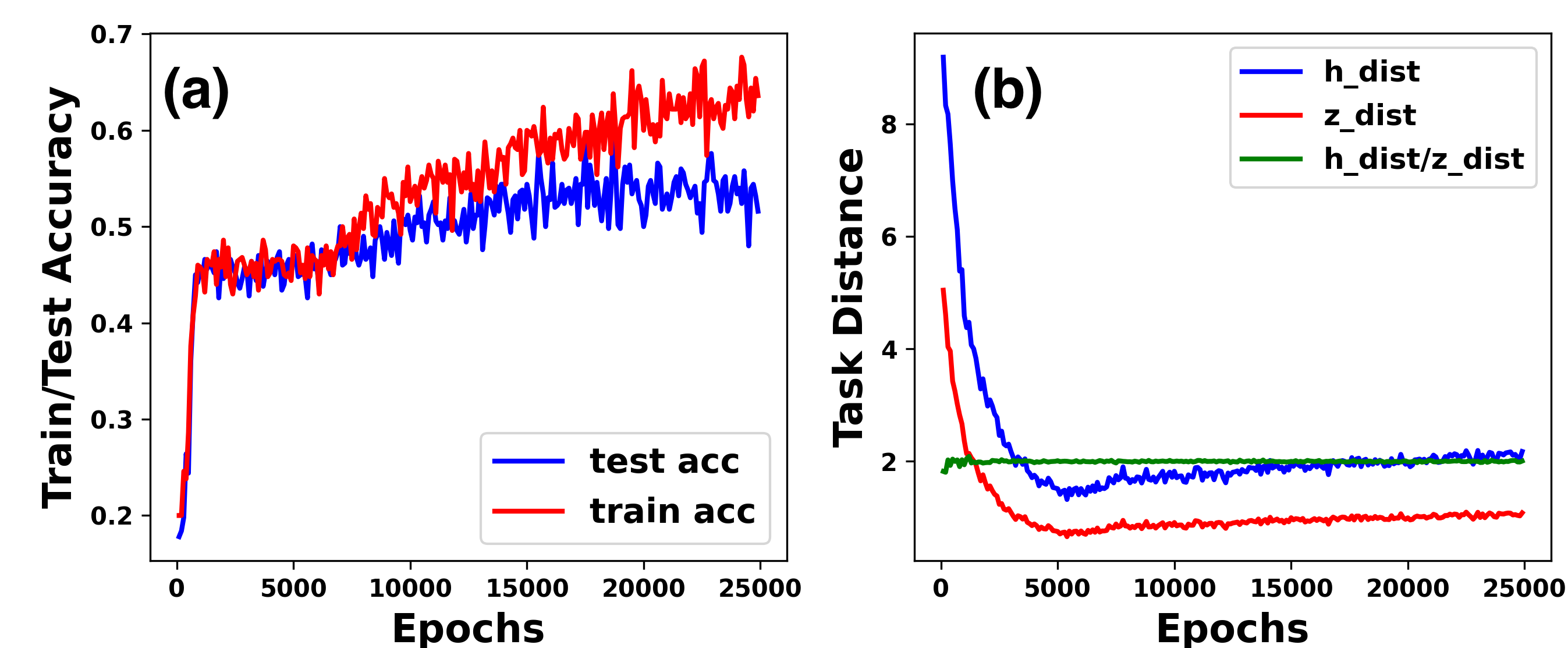}
    \caption{(a) Shows accuracy of black-box model(MANN) on 5 way 1 shot classification task with the regularization function as used in MER Method 3 in Table 1. (b) Here, we can also observe that distance between task-summary statistics has increased as compared to figure 3b where its low value  }\label{fig:MANN-5shot}
\end{figure}

In $5$-\textit{way} $1$-\textit{shot} classification, random guessing will produce accuracy of about 20\%. Looking at the results, we can clearly see that each of the MER-MANN methods of regularization produce better test accuracy compared to the MANN model with induced memorization. One interesting note involves the ratio of task-summary distance $h_\theta$ to last-latent information. Since a low ratio indicates that the model is ignoring the training data to make predictions (and is thus memorizing), one might expect that the higher the ratio the better. However, that is not the case. In fact, there appears to be a "sweet-spot" ratio where information is utilized from both task-summary and task-latent, resulting in performance exceeding that of the baseline label-shuffled MANN model (see Figure \ref{fig:MANN-5shot}). \\

In $10$-\textit{way} $1$-\textit{shot} classification, random guessing will produce accuracy around 10\%. Due to the significantly harder classification problem, we found this set of experiments difficult to optimize. In Table \ref{tbl:mer_mann_eta}, we detail multiple possible values of $eta$ that we tried in order to find the optimal ratio of information. Indeed, we found that $eta=4$ provided optimal results, which turn out to be slightly worse that that of the baseline label-shuffled MANN experiment.
\begin{table}[!hbt]\footnotesize 
\begin{center}
\caption{Validation accuracy of MANN models with our regularization, \textit{Meta Exclusiveness Regularization (MER)}, on Omniglot datasets with various different values for $\eta$.}
\label{tbl:mer_mann_eta}
\begin{tabular}{|l|*{2}{m{2.5cm}|}m{2.5cm}} \hline
 &  \multicolumn{2}{c|}{\textbf{\textit{10}}-way  \textbf{\textit{1}}-shot Classification} \\ \hhline{>{\arrayrulecolor{white}}-*{2}{>{\arrayrulecolor{black}}-}}
\multirow{-2}{*}{Desired Ratio ($\eta$)}& Test Accuracy \% & $h_{\theta} / z_{\theta}$ Ratio \\ \hline
 &  & \\
 \multirow{-2}{*}{2} & \multirow{-2}{*}{12.8}& \multirow{-2}{*}{2.0}\\ \hline
 &  & \\
\multirow{-2}{*}{4} & \multirow{-2}{*}{26}& \multirow{-2}{*}{4.0}\\ \hline
 &  & \\
\multirow{-2}{*}{6} & \multirow{-2}{*}{25}& \multirow{-2}{*}{6.0}\\ \hline
 &  & \\
 \multirow{-2}{*}{8} & \multirow{-2}{*}{24}& \multirow{-2}{*}{8.0}\\ \hline
 &  & \\
 \multirow{-2}{*}{10} & \multirow{-2}{*}{23.5}& \multirow{-2}{*}{10.0}\\ \hline
\end{tabular}
\end{center}
\end{table}

\subsection{MER-MAML}

\begin{figure}[!ht]
\centering
    \begin{subfigure}{0.45\linewidth}
    \centering
    \includegraphics[width=\linewidth,height=1.6in]{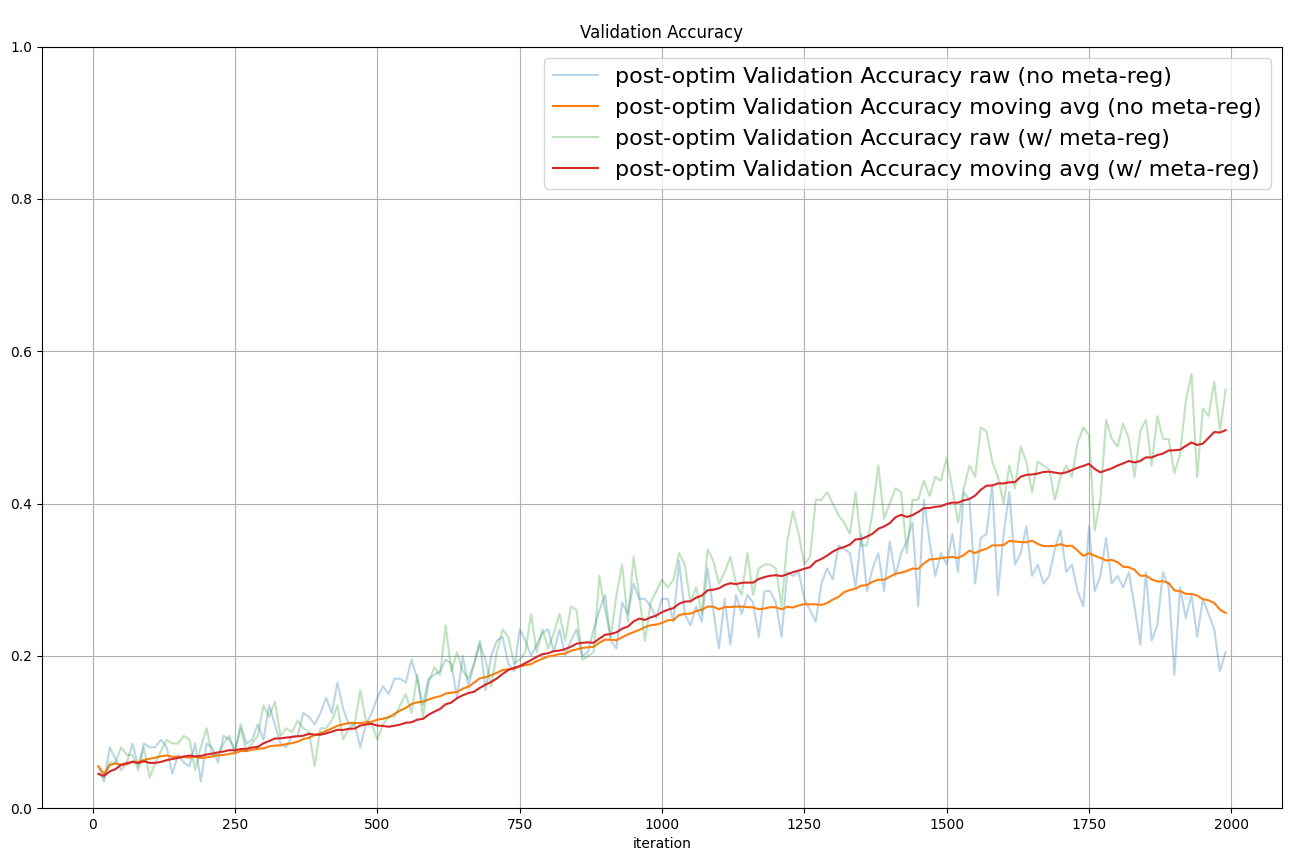}
    \caption{Post-optimization validation accuracy of MAML with and without \textit{Mutual Exclusiveness Regularization}}\label{fig:pre_post_optim_acc_maml_metareg}
    \end{subfigure}
    \hspace{0.1in}
    \centering
    \begin{subfigure}{0.45\linewidth}
    \vspace{0.2in}
    \includegraphics[width=\linewidth,height=1.6in]{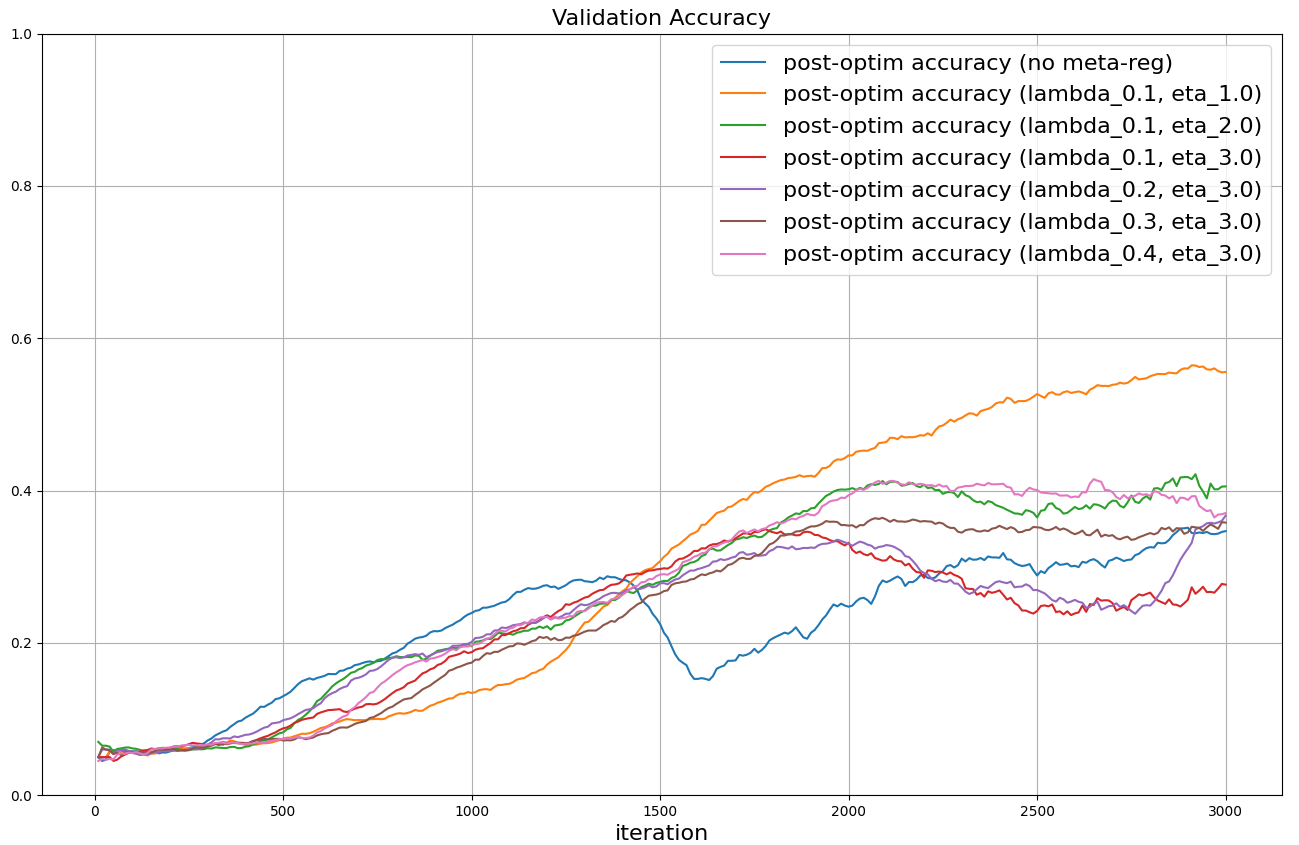}
    \caption{Validation accuracy of MAML with our regularization, \textit{Meta Exclusiveness Regularization (MER)}, on omniglot dataset for various combinations of $\lambda$ and $\eta$}\label{fig:metareg_omniglot_ablation}
    \end{subfigure}
    \caption{Validation accuracy of MAML and MER-MAML.}\label{fig:validation_acc_maml_mer_maml}
\end{figure}

For Omniglot dataset, memorization problem in MAML is recognized when the pre-optimization accuracy starts to approach post-optimization accuracy. This meta-overfitting also results in low \textit{meta-validation} accuracy as outlined in Fig \ref{fig:fig3-ver2}c. Using our meta exclusiveness regularization (MER) method with MAML, we are able to improve the \textit{meta-validation} accuracy. This is illustrated in Fig \ref{fig:pre_post_optim_acc_maml_metareg} which shows the contrast between MAML (\textit{without MER}) and MER-MAML. As seen in the plot, without MER, the validation accuracy first increases and then starts to decay as the training progresses and model starts to memorize \textit{meta-training} samples. MER-MAML alleviates this problem as evident from the graph, where we see a gradual improvement in \textit{validation accuracy}.  

We conduct various experiments on the Omniglot dataset classification problem by varying the regularization factor $\lambda$ and task-to-data encoding ratio $\eta$. In all our experiments, we find that our method, MER-MAML, improves accuracy over the model without any meta-regularization (MAML). We achieve the best results with $\lambda$ = $0.1$ and $\eta$ = $1.0$, which improves the meta-testing accuracy by $\sim 36\%$. These results are summarized in Table \ref{tab:ablation_mer_maml}. \\

A plot of validation accuracy for various $\lambda$ and $\eta$ are shown in Fig \ref{fig:metareg_omniglot_ablation} which shows a trend that MER improves validation accuracy as training progresses. 

\begin{table}[!ht]
    \begin{center}
    
      \caption{Testing accuracy of MAML with our regularization, \textit{Meta Exclusiveness Regularization (MER)}, on omniglot dataset for various combinations of $\lambda$ and $\eta$}
      \label{tab:ablation_mer_maml}
      
    \setlength{\tabcolsep}{0.5em}
    \def\arraystretch{1.5}
    \begin{tabular}{|l|l|l|l|l|l|l|}
    \toprule
    \textbf{\textit{n}}-way & \textbf{\textit{k}}-shot & Method   & \multicolumn{1}{l|}{$\lambda$} & \multicolumn{1}{l|}{$\eta$} & \multicolumn{1}{l|}{accuracy} & \multicolumn{1}{l|}{std. deviation} \\ 
    \midrule
    20 & 1 & MAML     & \multicolumn{1}{l|}{-}      & \multicolumn{1}{l|}{-}   & 0.22516668                    & 0.16212127                          \\ \hline
    20 & 1 & MER-MAML & 0.1                         & 1                        & \textbf{0.58416665}           & \textbf{0.13928139}                 \\ \hline
    20 & 1 & MER-MAML & 0.1                         & 2                        & 0.41899998                    & 0.16919309                          \\ \hline
    20 & 1 & MER-MAML & 0.1                         & 3                        & 0.3726667                     & 0.24226156                          \\ \hline
    20 & 1 & MER-MAML & 0.2                         & 3                        & 0.3465833                     & 0.17981702                          \\ \hline
    20 & 1 & MER-MAML & 0.3                         & 3                        & 0.24208333                    & 0.20874585                          \\ \hline
    20 & 1 & MER-MAML & 0.4                         & 3                        & 0.39549997                    & 0.25433853                          \\ \bottomrule
    \end{tabular}
\end{center}
\end{table}

We observe that the MAML architecture that consists of a data encoder and a task solver leads to a higher accuracy as compared to a simpler model that does not have a separate data encoder. In this similar model, we directly used the distance between $D_i^{tri}$ to normalize the distance between task-specific parameters $\phi_i$. Table \ref{tab:ablation_mer_maml_simpler} summarizes the testing accuracy of a \textit{simpler} model, and as is evident from the table, the model achieves lower accuracy overall. MER-MAML does improve the accuracy on Omniglot dataset even with a simpler model and achieves the best result for $\lambda=1$, $\eta=4$ but is lower then the complex model as shown in Table \ref{tab:ablation_mer_maml} where we have separately learned task embeddings to normalize task parameters.

\begin{table}[!ht]
\begin{center}
\caption{Testing accuracy of a simpler MER-MAML architecture on omniglot dataset for various combinations of $\lambda$ and $\eta$}
\label{tab:ablation_mer_maml_simpler}\setlength{\tabcolsep}{0.5em}
\def\arraystretch{1.5}
\begin{tabular}{|l|l|l|l|l|l|l|}
\toprule
\textbf{\textit{n}}-way & \textbf{\textit{k}}-shot & Method & $\lambda$ & $\eta$ & accuracy       & std. deviation           \\ \midrule
20             & 1   & \textit{simple} MAML   &  -  &  -  & 0.12           & 0.07          \\ \hline
20             & 1  & \textit{simple} MER-MAML   & 1  & 1  & 0.140 & 0.07 \\ \hline
20             & 1  & \textit{simple} MER-MAML    & 1  & 3  & 0.215          & 0.08          \\ \hline
20             & 1   & \textit{simple} MER-MAML   & 1  & 4  & \textbf{0.2787}         & 0.09          \\ \hline
20             & 1    & \textit{simple} MER-MAML  & 1  & 5  & 0.2717         & 0.08          \\ \bottomrule
\end{tabular}
\end{center}
\end{table}

\begin{figure}[!htbp]
\centering
    \centering
    \includegraphics[width=0.55\linewidth]{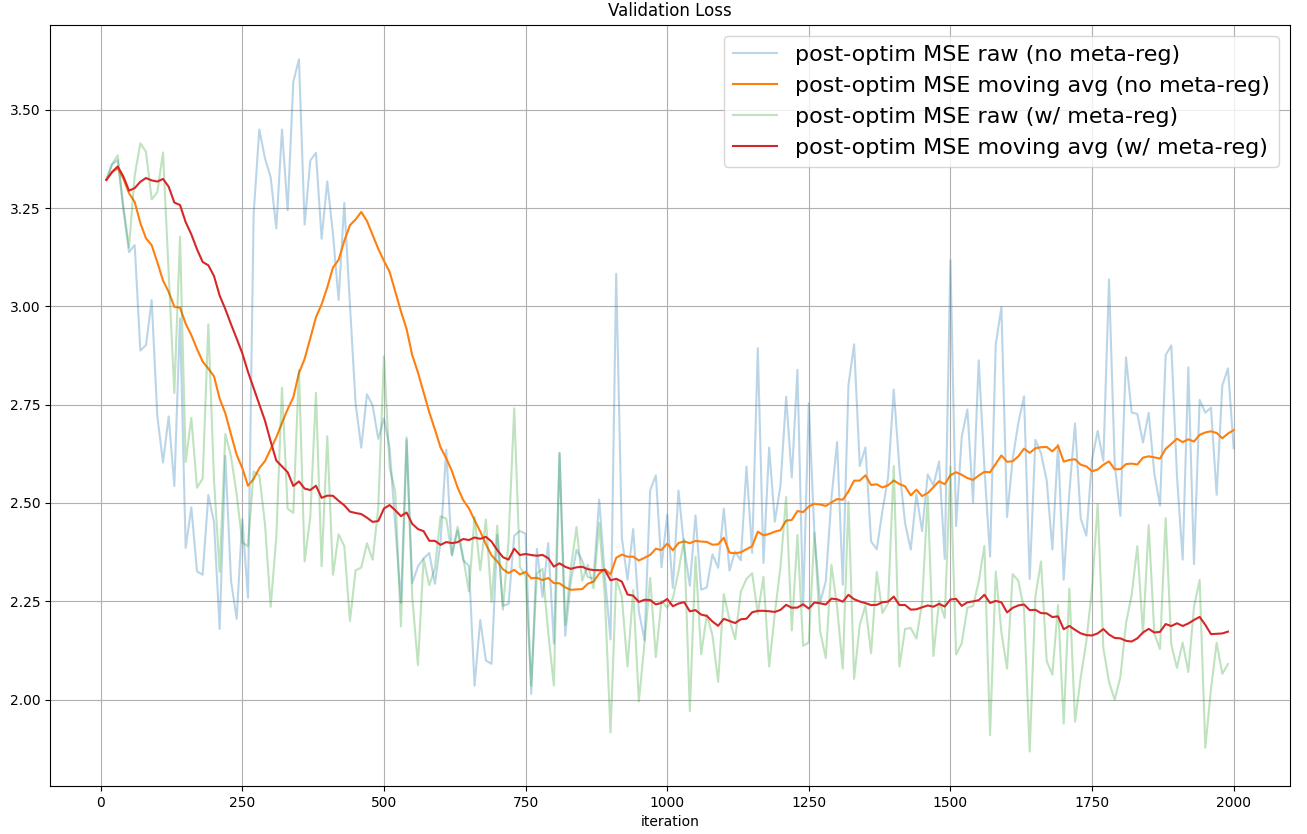}
\caption{MER improves the model performance on pose regression task.}\label{fig:metareg_maml_pose_mse}
\end{figure}

Some preliminary work with \textit{Pose} dataset for regression task also seems to improve the model performance. Fig \ref{fig:metareg_maml_pose_mse} shows \textit{validation} MSE of MER-MAML on pose regression task. We notice that our regularization method decreases the \textit{test} MSE on pose regression task by $0.63$ $rad^2$. 

\section{Conclusion}
In this paper, we overview the memorization problem in meta-learning training. Specifically, we started by formalize the meta-learning problem in relation to the canonical single task machine learning problem. With this view, we show that memorization can be viewed as a form of task-overfitting introduced by the more sophisticated structure of the meta-training procedure. Naturally, we illustrate the need for mutually-exclusive tasks, and describe how to induce memorization in the $N$-\textit{way} $K$-\textit{shot} image classification problem extensively used in this paper. \\

We show that our proposed regularization method, \textit{Meta Exclusiveness Regularization (MER)}, improves model performance in various frameworks including black-box and optimization-based meta-learning approaches. For the \textit{classification} problem, we explore the Omniglot dataset on which our method consistently improves \textit{validation} and \textit{test} accuracy. Furthermore, our preliminary experiments also show an improved \textit{test} MSE in a regression task for which we used \textit{Pose} dataset. \\

By enforcing \textit{Mutual-Exclusiveness}, we are able to avoid meta-overfitting during meta-learning. From our experiments, we found that the \textit{mutual-exclusiveness} between tasks (i.e. distance between task-specific parameters) should be proportional to the distance between data-points. The proportionality ratio was kept a constant during all of our experiments and the best one was empirically found. In our future work, we plan to learn this hyper-parameter, as well as learn a better representation of task-latent space.  


\clearpage

%
%
\bibliographystyle{splncs04}
\bibliography{metareg}
\end{document}